# ChatGPT-Like Large-Scale Foundation Models for Prognostics and Health Management: A Survey and Roadmaps

*Yan-Fu Li [a,b*], Huan Wang [a,b*] and Muxia Sun [a,b]*

*[a]Department of Industrial Engineering, Tsinghua University, Beijing, 100084, China*

*[b]Institute for Quality and Reliability, Tsinghua University, Beijing, 100084, China*

*Abstract*—**Prognostics and health management (PHM) technology plays a critical role in industrial production and equipment maintenance by identifying and predicting possible equipment failures and damages, thereby allowing necessary maintenance measures to be taken to enhance equipment service life and reliability while reducing production costs and downtime. In recent years, PHM technology based on artificial intelligence (AI) has made remarkable achievements in the context of the industrial IoT and big data, and it is widely used in various industries, such as railway, energy, and aviation, for condition monitoring, fault prediction, and health management. The emergence of large-scale foundation models (LSF-Models) such as ChatGPT and DALLE-E marks the entry of AI into a new era of AI-2.0 from AI-1.0, where deep models have rapidly evolved from a research paradigm of single-modal, single-task, and limited-data to a multi-modal, multi-task, massive data, and super-large model paradigm. ChatGPT represents a landmark achievement in this research paradigm, offering hope for general artificial intelligence due to its highly intelligent natural language understanding ability. However, the PHM field lacks a consensus on how to respond to this significant change in the AI field, and a systematic review and roadmap is required to elucidate future development directions. To fill this gap, this paper systematically expounds on the key components and latest developments of LSF-Models. Then, we systematically answered how to build the LSF-Model applicable to PHM tasks and outlined the challenges and future development roadmaps for this research paradigm.**

This work was supported by the National Natural Science Foundation of China under a key project Grant 71731008, and the Beijing Municipal Natural Science Foundation-Rail Transit Joint Research Program (L191022).

*Corresponding author: Yan-Fu Li, E-mail: liyanfu@tsinghua.edu.cn, Huan Wang, E-mail: huan-wan21@mails.tsinghua.edu.cn

# Contents

# 1. Introduction

Prognostics and health management (PHM) is a crucial technology for ensuring the safe and reliable operation of industrial equipment [1, 2]. By monitoring and managing equipment comprehensively, PHM reduces the probability of equipment failure and minimizes production downtime, thereby improving equipment reliability and production efficiency and creating significant economic benefits for enterprises [3, 4]. In industrial production practice, PHM has three core tasks: fault detection (anomaly detection) [5-7], fault diagnosis [8-11], and remaining useful life (RUL) estimation [12-14]. Anomaly detection aims to timely identify abnormal activities and states of equipment, while fault diagnosis aims to determine the cause and location of equipment faults. RUL estimation, on the other hand, predicts the time when equipment may fail in the future. These three tasks work together from different perspectives to ensure the safe operation of equipment. As industrial equipment becomes more sophisticated and complex, and the volume of operational monitoring data grows, automation of industrial data analysis, equipment status monitoring, and health management is necessary [15]. This automation can significantly reduce the maintenance costs of industrial assets, improve the efficiency and accuracy of equipment status identification and fault prediction, and enhance the reliability and safety of equipment operation.

In recent years, significant progress has been made in the field of PHM with the advancement of machine learning and deep learning [16-18] technologies, enabling automated industrial equipment condition monitoring and fault prediction, and greatly improving the intelligent level of industrial asset maintenance. Since the early 20th century, machine learning technology has played a crucial role in realizing intelligent identification and decision-making in PHM [19-21]. The PHM models based on machine learning mainly comprise two core components: feature engineering and machine learning models. Feature engineering leverages statistical analysis and signal analysis techniques [22-26] to extract health-related feature information from industrial monitoring data. Machine learning models use various prediction and identification models, such as support vector machines (SVM) [27-29] and $K$-nearest neighbor (KNN) [30-32], to achieve intelligent decision-making. This research paradigm has enabled PHM to achieve preliminary automation and reduced the demand for manual labor in industrial equipment maintenance. However, despite the progress made, manual feature engineering is still required, limiting the ability of PHM to cope with large-scale data. The limited learning capacity of machine learning models makes it difficult for this paradigm to adapt to the challenges posed by the big data era.

Since 2012, deep learning technology [33-35] has revolutionized the paradigm of various research fields with its powerful data analysis, feature extraction, and intelligent decision-making capabilities. Deep learning realizes automatic feature extraction and pattern recognition of complex data by establishing a multi-level neural network structure, which can automatically process high-dimensional, nonlinear, and large amounts of data, and has adaptive and generalization capabilities. Therefore, deep learning has become a mainstream tool in the PHM field [17, 18, 36], continuously improving the automation and intelligence of industrial asset maintenance. Various deep network models have been proposed for different PHM applications and tasks, such as autoencoders [37-39], convolutional neural networks (CNNs) [40-49], and recurrent neural networks (RNNs) [50-52]. Autoencoders achieve unsupervised representation learning through data compression or reconstruction, performing well in tasks such as data noise reduction, dimensionality reduction, and anomaly detection [53]. CNNs, based on convolution theory, achieve efficient spatio-temporal feature extraction by weight sharing and hierarchical learning, making them suitable for health monitoring, fault prediction and diagnosis, and RUL prediction of industrial equipment [54]. RNNs excel at encoding long-distance timing features, making them ideal for analyzing and processing various timing signals [55]. Therefore, RNNs are widely used in various industrial PHM applications. Deep learning technology has significantly reduced the need for manual labor in industrial PHM applications by building an end-to-end intelligent decision-making model [8]. However, existing deep learning models still have limitations in multi-task, generalization, and cognitive abilities. Therefore, breaking through these limitations to realize a comprehensive multi-task intelligence model with high generalization and cognitive ability is an urgent problem that requires a solution.

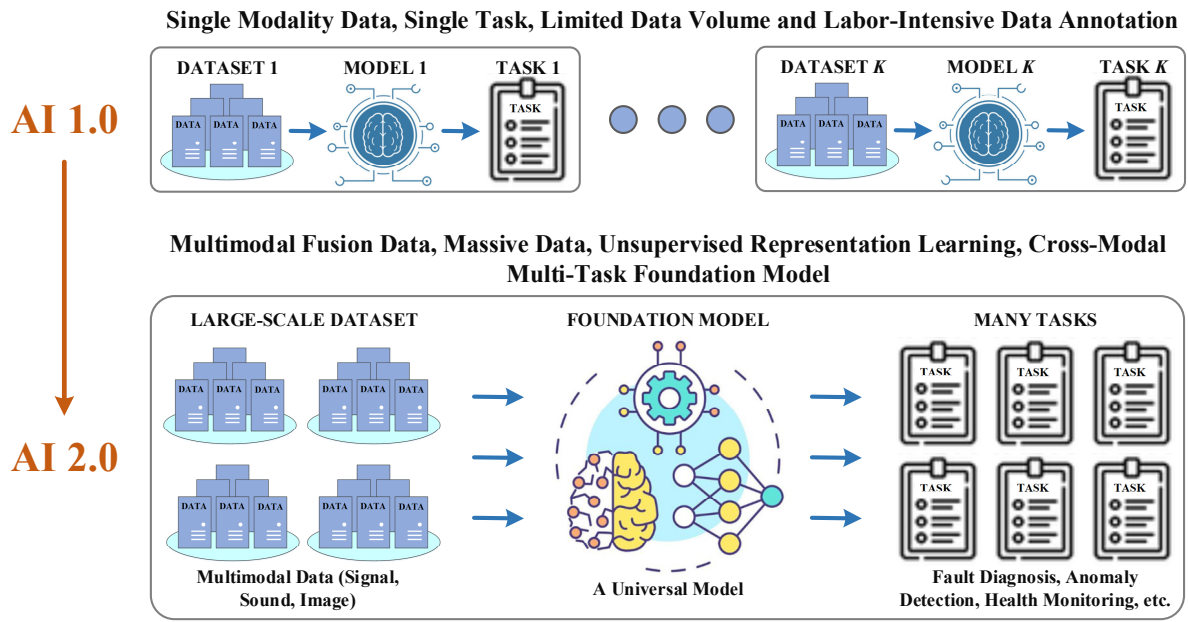

**Fig. 1**. From AI-1.0 to AI-2.0, the research paradigm of deep learning has undergone tremendous changes.

In the past two years, large-scale Foundation models (LSF-Models) [56, 57], such as GPT-3 [58, 59] and ChatGPT [60, 61], have demonstrated highly intelligent natural language understanding capabilities with fluent text dialogue. Large-scale multi-modal text and image understanding models, such as GPT-4 [62], DALL-E-2 [63], and segment anything model (SAM) [64], further demonstrate this research paradigm's extraordinary achievements in multi-modal dialogue, image generation and segmentation. The AI-based deep model has developed rapidly from a single-modal, single-task, limited-data research paradigm (AI-1.0) to a multi-modal, multi-task, massive data and super-large model research paradigm (AI-2.0). **Fig. 1** clearly shows the difference between these two research paradigms. The core of AI-2.0 is the LSF-Model with cross-domain knowledge, which can understand the general notion of data and achieve Zero-shot generalization on unseen data without additional training [64]. The realization of this model is mainly based on the following three key components, the potent feature extraction model [65-68], the unsupervised representation learning algorithm [69-71], and the multi-modal fusion algorithm [72, 73]. Moreover, extensive unlabeled or labeled multi-modal data is a prerequisite for this development. However, how to build the LSF-Model with cross-domain knowledge in the PHM field is still unknown, and there is still a lack of sufficient research and analysis on how to develop feature extraction, representation learning, and multi-modal fusion algorithms suitable for the PHM field. Moreover, how the PHM field responds to this enormous change in the AI field is still inconclusive, and there are lack of systematic literature review, as well as roadmaps for future research direction. In order to fill this gap, this paper first systematically presents the key components and cutting-edge progress of LSF-Models. Then, we systematically answer how to build effective LSF-Model suitable for the PHM field. We also elaborated on the challenges this research paradigm will face and the roadmaps for future development.

Specifically, the main work of this survey is summarized as follows:

1) This paper provides a comprehensive review of the three key components and the respective research progress of LSF-Models.

2) Drawing from the actual circumstances in the PHM field, this paper systematically analyzes and answers how to build effective LSF-Model suitable for industrial PHM applications.

3) This paper discusses the roadmaps of LSF-Model research in the PHM field and analyzes in detail the challenges and solutions faced by this research paradigm.

The rest of this paper is organized as follows. Section 2 focuses on the key components of LSF-Models, including Transformer, self-supervised learning, and multi-modal fusion. Section 3 reviews the research progress of LSF-Models in natural language processing and computer vision. Section 4 systematically answers how to implement LSF-Models for PHM, including research status, existing problems, and solutions. Section 5 comprehensively discusses the challenges of LSF-Model research in the PHM field and its future roadmaps. Conclusions are presented in Section 6.

## 2. Key Components of Large-Scale Foundation Models

LSF-Models are a class of large-scale deep neural network models consisting of billions of parameters [57]. These models are trained on massive amounts of data to enable them to capture the complex relationships and general notions of the data, thus possessing cross-task and cross-domain zero-shot generalization capabilities. The development of LSF-Models has benefited from the advancement of various technologies, including the improvement of computing hardware, the availability of big data, the development of representation learning, the improvement of model architecture, and the advancement of multi-modal data fusion algorithms. The improvement of computing hardware and the availability of big data are the basic conditions for LSF-Models [74]. This is mainly due to the development of computing hardware and the Internet in recent years, which respectively provide the hardware foundation and data foundation for the realization of LSF-Models. Additionally, the development of algorithms and neural networks has played an inseparable role in the development of LSF-Models. For instance, the proposal of transformer architecture provides powerful feature extraction capabilities for the foundation model [66, 68]. Self-supervised learning has facilitated the development of a powerful unsupervised feature representation capability for the foundation model [69, 70]. Furthermore, multi-modal fusion algorithms have equipped the foundation model with the ability to interact across modalities [72]. In the subsequent section, we discuss and analyze these three aspects in detail.

## 2.1 Transformer-Based Feature Extraction

### 2.1.1 Network Architecture of Transformer

Transformer [75] is a powerful network model based on the self-attention mechanism, which was first used in applications such as sequence modeling and natural language processing (NLP) [76]. **Fig. 2** shows the basic architecture of the Transformer, which is a typical encoder-decoder architecture, where the encoder and decoder are composed of multiple transformer blocks. The encoder encodes the input sequence into a hidden vector representation, and the decoder synthesizes the contextual information of the hidden vector to generate sequence information. Each Transformer block contains a multi-head attention (MHA) mechanism, a feed-forward neural network (FFNN), residual connections [77], and layer normalizations [78]. It is described in detail below.

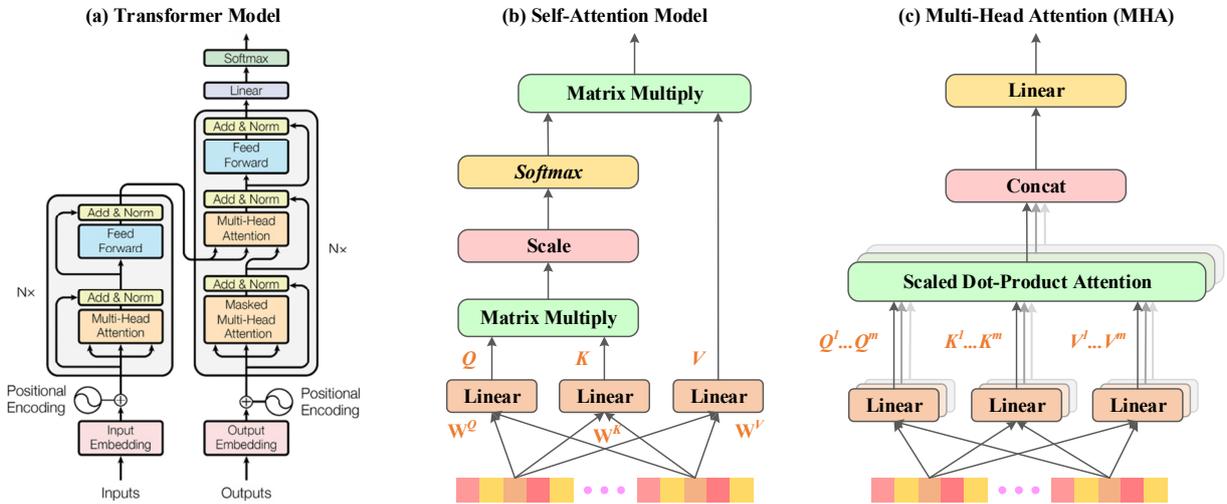

**Fig. 2**. Architectural details of the Transformer model, self-attention model, and multi-head attention (image from [42, 75]).

**Self-Attention Mechanism.** Assuming that the input feature sequence of the self-attention mechanism is represented as $X = [x_1, x_2, \cdots, x_k]$, $X$ is encoded into $Q$ (query vector), $K$ (key vector), and $V$ (value vector) by three sets of linear transformation layers. The learnable weights of these three sets of linear transformation layers are denoted as $\mathbf{W}^Q, \mathbf{W}^K, \mathbf{W}^V$, respectively. Based on the obtained $Q$, $K$, and $V$, the self-attention mechanism can be expressed as:

$$Att\left(Q,K,V\right)=Softmax\left(\frac{\left(X\mathbf{W}^Q\right)\left(X\mathbf{W}^K\right)^T}{\sqrt{d}}\right)\left(X\mathbf{W}^V\right)=Softmax\left(\frac{QK^T}{\sqrt{d}}\right)V \qquad (1)$$

where $\left(X\mathbf{W}^Q\right)\left(X\mathbf{W}^K\right)^T$ is to calculate the similarity between $Q$ and $K$, and then input the scale-transformed similarity into Softmax to obtain the attention weight, and finally perform matrix multiplication between the weight and $V$ to obtain the final output. In addition, d represents the dimensions of $Q$ and $K$, and $1/\sqrt{d}$ is a scale factor to avoid excessive feature values. The advantage of the self-attention mechanism is that it can consider the relationship between all positions in the sequence, not just the relationship between adjacent positions. This global relational modeling capability is essential for many NLP tasks, such as machine translation, text generation, and language understanding.

**Multi-Head Attention Mechanism.** MHA is an extended form of self-attention mechanism, which can improve the model's ability to model different information. MHA encodes the input vector into multiple subspaces, calculates the self-attention mechanism in each subspace, and finally stitches the attention results of different subspaces to obtain the final attention result. This enables the model to encode global contextual features of input features from different perspectives. MHA can be expressed as:

$$\begin{aligned}S=MHA\left(X\right)&=Concat\left(head^1,head^2,\cdots head^H\right)\mathbf{W}^H,\\ where\quad head^H&=Att\left(X\mathbf{W}_H^Q,X\mathbf{W}_H^K,X\mathbf{W}_H^V\right)\end{aligned} \qquad (2)$$

where $X$ and $S$ denote the input and output of the MHA, respectively; $H$ represents the number of attention heads; $Concat\left(\cdot\right)$ means splicing the output results of $H$ attention heads, and $head^H$ means an attention head, that is, a self-Attention mechanism module; in the MHA, there are $H$ parallel attention heads. $\mathbf{W}^H$ denotes the weight of a fully connected layer, which is used to fuse the output weights of multiple attention heads.

**Feed-Forward Neural Network.** In the Transformer block, following the MHA is the FFNN, which consists of two linear transformation layers and a nonlinear activation function, which can be expressed as:

$$FFN\left(S\right)=\mathbf{W}_2\rho\left(\mathbf{W}_1S\right) \qquad (3)$$

where $\mathbf{W}_1$ and $\mathbf{W}_2$ are the parameters of the two linear transformation layers, respectively, and $\rho\left(\cdot\right)$ represents the nonlinear activation function, such as ReLU and GELU [79]. Its primary function is to enable more complex mapping and transformation of the output from the self-attention mechanism, thereby enhancing the processing of intricate features and semantic information.

**Layer Normalization and Residual Connection.** Layer normalization is utilized to enable the network to adapt to input data of differing scales and distributions. Specifically, through the normalization of output from each neuron, a similar data distribution is achieved across each layer, thus facilitating network optimization and training. Residual connections are implemented to enhance the transfer of gradient information throughout the network. As such, residual connection and layer normalization are added following MHA, which can be expressed as:

$$Z=LayerNorm\left(X+S\right), \qquad (4)$$

$$LayerNorm\left(X+S\right)=\frac{\left(X+S\right)-E\left(X+S\right)}{\sqrt{Var\left(X+S\right)+\varepsilon}}, \qquad (5)$$

where $E(\cdot)$ and $Var(\cdot)$ calculate the mean and variance based on each feature signal.

### 2.1.2 Research Progress of Transformer

The groundbreaking architecture design of the Transformer enables it to obtain excellent feature extraction performance, thereby attracting scholars to improve and optimize its key components continuously. The attention mechanism is the core component of Transformer, and its main improvement directions are 1) Sparse attention [80-82]; 2) Linearized attention [83-86]; 3) Optimizing the MHA mechanism [87-91]; 4) Alternatives to the attention mechanism [92-94]. Sparse attention is dedicated to introducing sparsity bias into the attention mechanism. Li et al. [81] proposed the sparse adaptive connection method in the self-attention architecture, which significantly reduces the computational complexity of attention. Linearized attention is dedicated to optimizing the feature interaction of self-attention to achieve linear complexity. Katharopoulos et al. [86] formulated self-attention as a linear dot product of kernel feature maps and utilized the matrix product's associativity to reduce computational complexity. The optimization of MHA focuses on enabling different attention heads to capture different valuable features adequately. Li et al. [91] introduced disagreement regularization in Transformer to make different attention heads have diversity and obtain valuable information from different subspaces. Alternatives to the attention mechanism are dedicated to finding new schemes to achieve a faster and more efficient global information interaction mechanism. For example, Rao et al. [94] have substituted the self-attention layer with a global filter layer based on Fast Fourier transform to capture global information at the frequency domain level.

In addition, scholars have conducted a considerable amount of work on improving and optimizing the activation functions [95, 96] and FFNNs [97-99] of the Transformer. In addition to research on the core components within the Transformer, optimizing the overall architecture of the Transformer is also a research focus. For example, Wu et al. [100] proposed a lightweight Lite Transformer that can be deployed on edge computing devices to execute NLP applications. Rae et al. [101] proposed a new sequence model: Compressive Transformer, which compresses past memories for long-range sequence learning. Liu et al. [102] introduced the idea of hierarchical feature learning in the Transformer architecture to obtain richer representations.

With the proposal of Vision Transformer (ViT) [103], it has gradually been introduced into image processing [66], video analysis [104, 105], and other related fields, which have also demonstrated excellent performance in these fields. To this end, various Vision Transformer architectures have sprung up [66, 68]. For example, Liu et al. [106] proposed a novel layered Transformer architecture based on shifted windows, which can be widely adapted to various vision applications. Huang et al. [107] further adopted the spatial shuffle operation to achieve cross-Window feature interaction, thereby improving the representation ability of the Transformer architecture in vision applications. Hu et al. [108] proposed a U-Net-like Swin Transformer model for visual image segmentation applications. The Transformer model has developed into a super-large system, and a large amount of research work has emerged. Please refer to the review papers [66, 68, 109] for a detailed review and analysis.

## 2.2 Self-Supervised Learning-Based Feature Representation

Self-supervised learning [69-71] is a paradigm of unsupervised learning that discovers effective feature representations by producing supervisory signals from unlabeled data. Unlike conventional supervised learning, which demands abundant labeled data, self-supervised learning exploits the abundant information of unlabeled data, thus reducing the reliance on manually labeled data. Typically, self-supervised learning devises a pretext task to enable the model to acquire valuable representations during the pretext task's resolution. The flowchart of self-supervised learning is shown in **Fig. 3**. Self-supervised learning has many classical algorithms in the fields of computer vision (CV) and NLP, mainly including the following methods.

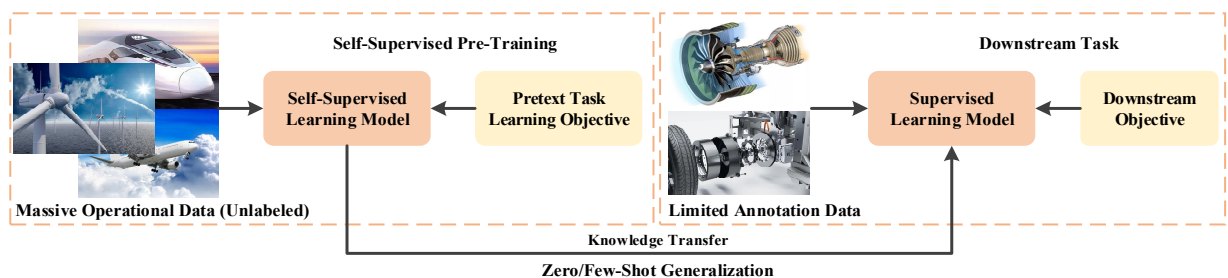

**Fig. 3**. Flowchart of the self-supervised learning algorithm, using the PHM application as an example.

**Masked Language Modeling (MLM).** The fundamental concept of this approach is to randomly mask specific words or characters in the input sequence and then prompt the model to predict the masked tokens [110, 111]. MLM is one of the most popular pre-training methods in the current NLP field. Many pre-trained models, including BERT [112] and RoBERTa [113], are designed based on the idea of MLM. In addition to MLM, there are also some related self-supervised learning methods, such as next-sentence prediction [114, 115] and back translation [116]. These techniques can aid in pre-training NLP applications to enhance the model's generalization ability and performance.

**Autoregressive Models**. This method is often used for pre-training in the NLP field, which predicts the next word based on a given context [117, 118]. For example, GPT [117] is a typical autoregressive-based method, and in this way, GPT can learn rich context-related information. As such, it has achieved notable success in NLP applications. XLNet [118] introduces a generalized autoregressive pre-training technique, which surpasses the constraints of BERT and accomplishes better contextual dependency learning. Oord et al. [119] proposed contrastive predictive coding that achieves self-supervised feature representation by predicting the future in the latent space, utilizing probabilistic contrastive loss to capture valuable information.

**Autoencoders**. Autoencoders are a classic representation learning algorithm, which consists of an encoder and a decoder [120-123]. The encoder maps the input data to a latent space, and the decoder reconstructs the input data from the latent space. These algorithms learn valuable feature representations from images by minimizing reconstruction errors. Lin et al. [120] proposed a Masked Autoencoder pre-training strategy that can handle both text and visual data, accomplishing multi-modal feature representation.

**Contrastive Learning**. This method aims to map similar samples to spaces that are close to each other and map dissimilar samples to spaces that are far away [124-126]. This approach helps models learn the differences and variations between data, thereby improving the model's generalization ability and performance [127]. Generally, the Contrastive Learning approach consists of two phases: constructing contrast pairs and training the model. In the contrast pair construction phase, techniques, such as random data augmentation, are commonly utilized to generate similar and dissimilar sample pairs. In the subsequent model training phase, a contrastive loss function is typically employed to minimize the distance between similar sample pairs and maximize the distance between dissimilar sample pairs. Contrastive learning can be applied in various domains, such as image [128, 129], audio [130, 131], natural language [132], etc. Specifically, within the domain of image processing, contrastive learning methods, such as SimCLR [128], SimCLRv2 [129], and MoCo [133], leverage CNNs and data augmentation techniques to learn the semantic features of images and enhance the performance of tasks like image classification and object detection.

**Image Augmentation-Based Self-Supervised Methods.** These methods mainly include rotation prediction, image colorization, image jigsaw puzzle, image inpainting, and image super-resolution. These methods are pretext tasks constructed based on the intrinsic characteristics of image data. Rotation prediction [134, 135] involves rotating the input image and predicting the rotation angle, enabling the model to learn the rotational invariance of the image. Image colorization [136, 137] involves adding appropriate colors to the input grayscale image, thereby facilitating the learning of valuable contextual information and enhancing the model's understanding of the data. The primary objective of the Image Jigsaw Puzzle [138-140] is to cut the image into pieces and reassemble them, enabling the model to learn the relationships between different parts of the image and improving its ability to learn image features. The main objective of Image Inpainting [141, 142] is to mask or delete certain regions in the image, enabling the model to learn how to infer and fill in the missing parts, thereby improving its ability to learn image features. Image Super-Resolution [143, 144] aims to generate high-resolution data from low-resolution data, allowing the model to learn detailed image information more effectively.

Self-supervised learning has the advantage of leveraging the inherent characteristics of data to enable the model to learn general and high-quality data features and potential relationships from a vast amount of unlabeled data. This facilitates the model's better understanding of the nature and laws of the data, significantly reducing the cost of manually annotated data. This characteristic also makes self-supervised learning suitable for large-scale datasets, and as a result, plays a crucial role in the research of LSF-Models. Consequently, a significant amount of research work related to self-supervised learning has emerged, and for a comprehensive review and analysis, please refer to the review papers [69-71, 125].

## 2.3 Multi-Modal Data Fusion

Multi-modal fusion [145-147] is a technique for integrating information from various modalities, such as text, images, audio, and video, in order to enhance model performance and generalization. The aim of multi-modal fusion is to extract more comprehensive feature representations by exploiting complementary information between multiple data sources, which in turn improves model performance in a wide range of applications, including sentiment analysis [148] and video question answering [149]. Furthermore, in the domain of LSF-Models, researchers strive to achieve interaction between multi-modal information and the comprehension of fundamental concepts, such as text-based image generation [63] and image captioning [150]. To date, multi-modal fusion approaches comprise the following methods.

**Early Fusion** [151, 152]. This method integrates information from different modalities at the input layer to obtain a comprehensive multi-modal representation that is subsequently input into a deep neural network for training and prediction.

**Late Fusion** [153-155]. This method independently extracts and processes features from different modalities in their respective neural networks and fuses the features at the output layer to obtain the final prediction result.

**Attention Fusion** [156-159]. This method utilizes attention mechanisms to weigh and fuse information from different modalities, enhancing the weight of important information and obtaining a more accurate multimodal representation and prediction result.

**Heterogeneous Fusion** [160-163]. This method combines information from different modalities on a heterogeneous graph to consider the characteristics and interrelationships of different modalities, thereby obtaining a more accurate multi-modal representation and prediction result.

**Prompt-Based Methods** [164-167]. This method introduces natural language prompts (Prompts) into the model to enhance its performance. In multi-modal fusion tasks, Prompts can guide the model to generate accurate cross-modal prediction results.

In LSF-Models research, multi-modal data fusion algorithms usually require the following properties. Multi-source: Multi-modal data fusion algorithms can fuse data from multiple sources simultaneously. Multi-level: Multi-modal data can be fused at different feature levels, improving data processing and analysis accuracy and efficiency. Diversity: Multi-modal data fusion algorithms can process data from multiple modalities, such as images, voice, text, etc. It can be seen that research on multi-modal fusion empowers models to comprehend various general concepts of the real world from multiple perspectives. Just as humans possess visual, auditory, tactile, and other sensory organs, general AI models will undoubtedly have strong multi-modal information perception abilities.

## 3. Progress on Large-Scale Foundation Models

In the preceding section, we deliberated on several key components that are imperative to construct LSF-Models. This section provides an extensive overview and discourse on the advancement of LSF-Models in the domains of NLP and CV, respectively, to exhibit the latest development trends and directions in these fields.

### 3.1 Large-Scale Foundation Models for Natural Language Processing

As a result of the progression of Internet technology, acquiring ultra-large-scale text data has become increasingly convenient, consequently enabling LSF-Models to attain remarkable advancements in the realm of NLP [168, 169]. Current large-scale language models can efficiently and accurately accomplish various NLP applications [170], such as language generation, language understanding, question-answering systems, machine translation, etc. Presently, a plethora of LSF-Models have surfaced in the NLP domain, and the principal models are delineated below.

**BERT Series Models.** Bidirectional Encoder Representations from Transformers (BERT) [112] is a language model based on the pre-trained bidirectional Transformer architecture. BERT employs pre-training to acquire universal language representation from the voluminous unlabeled text. The pre-training tasks employed by BERT include MLM and Next Sentence Prediction. Subsequently, Liu et al. [113] enhanced BERT by proposing the RoBERTa model, which implemented optimization strategies such as prolonged training time, increased training data, dynamic masks, and finer-grained batch size. Furthermore, Lan et al. [171] developed a lightweight variant of BERT (ALBERT) through techniques such as parameter sharing and embedding decomposition, effectively reducing the resource consumption and training duration of the model.

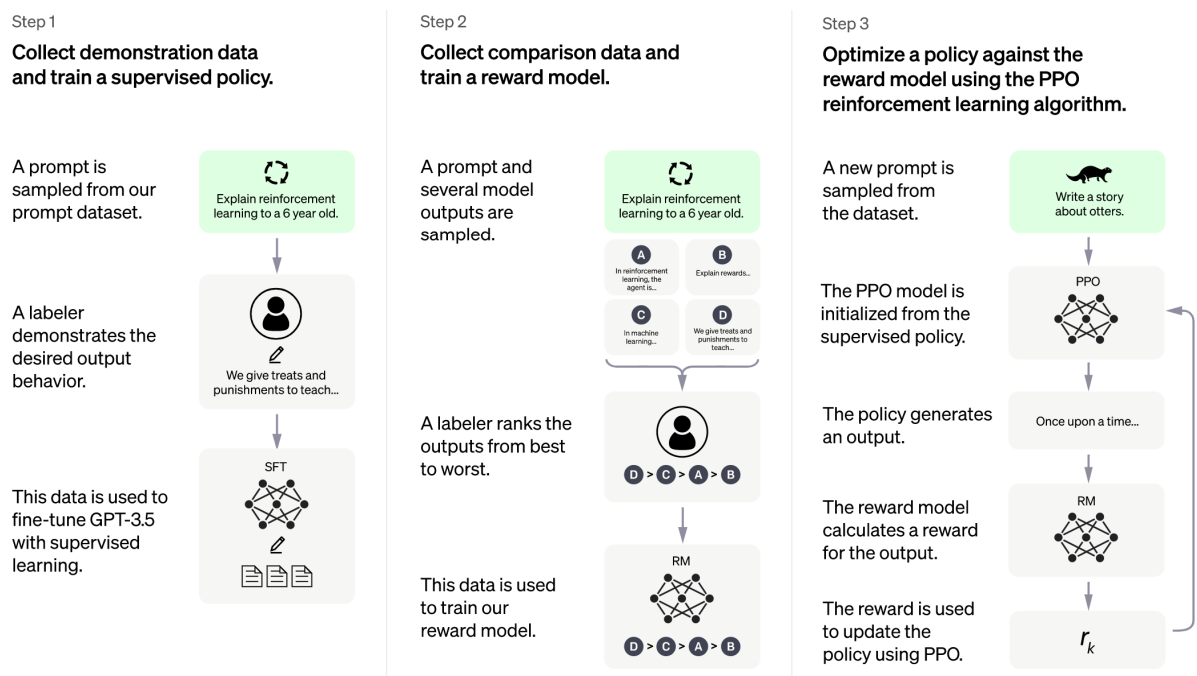

**Fig. 4**. Flowchart of ChatGPT's training algorithm, whose core is reinforcement learning based on human feedback (image from [60]).

**GPT Series Models.** The GPT series is a collection of Transformer-based pre-trained language models developed by OpenAI, consisting of GPT1.0 [172], GPT 2.0 [173], GPT 3.0 [58], GPT 3.5 (ChatGPT) [60], and the most recent model, GPT-4 [62]. The model utilizes the autoregressive approach to perform unsupervised pre-training on a massive corpus, enabling the learning of high-level representations of text and vocabulary, which facilitates fine-tuning and transfer learning in various downstream tasks. The primary characteristic of the GPT series model is the utilization of unlabeled text data for pre-training, combined with the self-attention mechanism and position encoding of the Transformer. Additionally, the adoption of the autoregressive model allows the model to process sequential information in text effectively. **Fig. 4** shows the three training steps of GPT 3.5 and their details. GPT 3.5 has demonstrated potent natural language understanding capabilities, and the latest model, GPT-4, integrates image understanding capabilities, resulting in highly intelligent multi-modal information processing capabilities.

**ERNIE Series Models.** The ERNIE series are pre-trained language models based on the Transformer architecture developed by Baidu. The series includes ERNIE 1.0 [174], ERNIE 2.0 [175], and ERNIE 3.0 [176]. These models adopt the training method of the autoregressive language model and MLM and incorporate two new pre-training tasks, namely text cloze and sentence rearrangement, to enhance the model's contextual understanding and sentence-level semantics. ERNIE expands the model's corpus and knowledge base by designing a new knowledge distillation and data enhancement strategy. Moreover, ERNIE 2.0 and ERNIE 3.0 adopt a knowledge distillation approach to transfer knowledge from a teacher model to a smaller student model, which improves the model's efficiency and allows it to be deployed on devices with limited computational resources.

**T5 Series Models.** The T5 series is a general-purpose text generation framework developed by Google [177]. It adopts the Transformer structure. Its core idea is to unify various NLP tasks into text-to-text conversion problems, simplifying pre-training and fine-tuning. The T5 model uses the Encoder-Decoder structure, where the Encoder is responsible for encoding the input text, and the Decoder is responsible for decoding the output text. The latest improved version Plan-T5 [178], can be used for almost any NLP task by fine-tuning on ultra-large-scale tasks and initially realizes the one model for all tasks.

In addition, numerous companies and research institutions have developed a plethora of high-quality large-scale language models. For instance, Meta Corporation recently unveiled its LLaMA model [179], which boasts a massive parameter volume of 65 billion. Similarly, Huawei has proposed a trillion-parameter language model [180] that exhibits exceptional performance on various Chinese NLP tasks. Another noteworthy model, Sparrow [181], developed by DeepMind, aims to align the model's responses more closely with the user's intentions and provide more accurate replies based on the searched content. Currently, large-scale language models are continually being refined and optimized. Please refer to the review papers [182, 183] for a detailed review and analysis.

## 3.2 Large-Scale Foundation Models for Computer Vision

Inspired by the resounding success of large-scale language models in the NLP field, researchers have explored the application of LSF-Models in the CV field [64, 184, 185]. Similarly, visual foundation models involve representation learning on large-scale image datasets to achieve cross-domain and high-level semantic understanding while enabling multi-tasking capabilities. Since gathering and analyzing massive image data is more challenging than text data, most traditional computer vision pre-training models rely on well-established classification architectures, such as ResNet [77], to train on the ImageNet dataset [74]. The pre-trained weights obtained from such models are then transferred to downstream tasks, improving performance on various visual tasks. However, despite ImageNet containing more than 14 million images, these conventional pre-trained models still fall short of possessing exceptional multi-task capabilities, few-shot generalization capabilities, and the ability to learn general image concepts. Recently, thanks to advancements in algorithms and the availability of vast amounts of image data, general-purpose large-scale visual models are being developed.

In 2023, Meta AI released the first large-scale general-purpose visual image segmentation model: Segment Anything Model (SAM) [64, 184]. They also released the largest image segmentation dataset, which contains 11 million images and 1 billion masks. SAM stands out for its remarkable ability to perform both interactive and automatic segmentation as a single model. Additionally, it exhibits a profound understanding of general concepts present in image data, enabling it to generate masks for any object in any image or video, even those that it did not encounter during training. This zero-shot generalization capability eliminates the need for domain-specific data collection for fine-tuning the model.

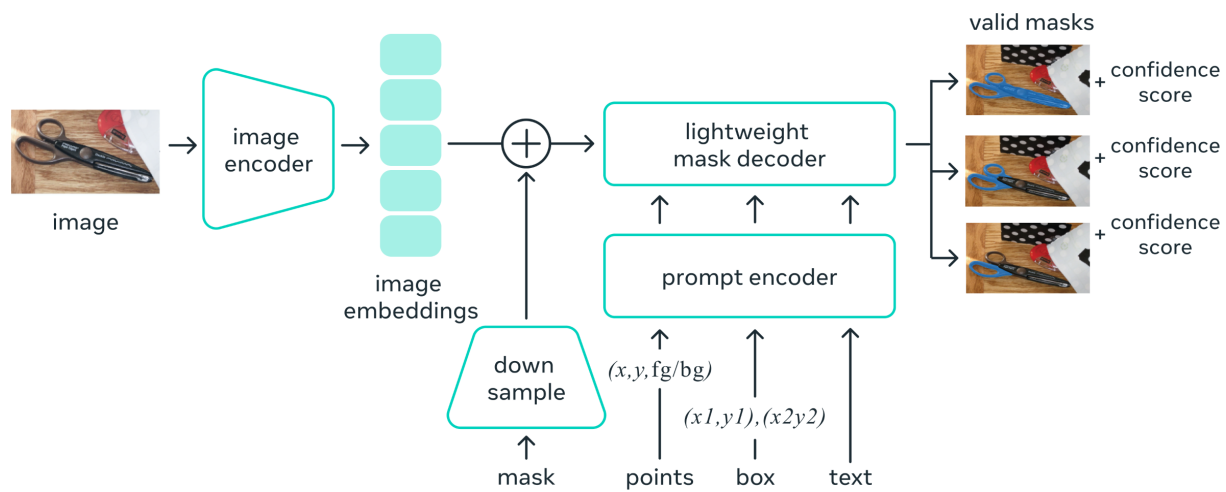

**Fig. 5**. The network architecture and algorithm flow chart of SAM, which can output segmentation results according to the input prompts (image from [64, 184]).

Contrastive Language-Image Pretraining (CLIP) [186] is a multi-modal pretraining model developed by OpenAI, which captures the joint representation of vision and language by simultaneously learning image and text information. CLIP employs contrastive learning as a training strategy to learn a joint representation of vision and language by comparing matching images and text (positive examples) with mismatched images and text (negative examples). Specifically, CLIP maximizes the similarity between matching images and text, while minimizing the similarity between unmatched images and text. The model was trained on a dataset of 400 million images and text collected from the Internet. It also enables zero-shot generalization to unseen data, enabling accurate image-to-text descriptions.

DALL·E [185] is an image generation model based on Transformer and GAN proposed by OpenAI. The core of the DALL·E model is a dual-stream Transformer architecture, one of which is used to process text input, and the other is used to process image generation. It can automatically understand objects, scenes, and situations described by natural human language and generate high-quality images. DALL·E 2 [63] is a further improved version, which can generate more realistic and high-resolution images.

Similarly, the vision-language representation learning model [187, 188], akin to the CLIP model [186], is undergoing rapid development intending to achieve cross-modal comprehension and interaction between images and natural language. At present, the mainstream models of this paradigm include ViT-BERT [189], BLIP [190], and ZERO [191], etc. These models are most designed based on the Transformer architecture, which obtains the cross-representation of vision and language through pre-training learning and performs fine-tuning in various tasks. These models have performed well in various applications of vision and language, laying a solid foundation for proper visual and language understanding.

## 4. Large-Scale Foundation Models for PHM

While deep learning models have achieved remarkable performance in PHM, their effectiveness is limited because they are typically trained and optimized for specific domains and tasks. As a result, in complex and open industrial scenarios, deep learning models may exhibit several limitations, such as limited generalization, multi-tasking, and cognitive abilities. Existing models may perform well in known scenarios, but are challenging to generalize effectively to unknown scenarios [192-194]. This lack of zero-shot generalization ability makes it difficult to cope with the complexity of actual industrial scenarios. Furthermore, existing deep models usually focus on a single task. However, industrial equipment has hundreds of core components, all requiring health monitoring and fault prediction. It is unrealistic to develop a corresponding deep model for each core component. Finally, the existing deep models have limited cognitive ability, and it is difficult to understand the nature and general concepts of industrial data, so they often output incomprehensible and wrong results.

The study of LSF-Models shows us an efficient solution that can successfully solve the above problems. As described in Section 3, existing LSF-Models, such as ChatGPT [62] and SAM [64], have demonstrated excellent data understanding, zero-shot generalization, and strong multi-task capabilities. Moreover, it also has certain advanced cognitive abilities to solve some reasoning tasks. Therefore, the success of LSF-Models marks the transformation of the research paradigm in the AI field from a single-modal, single-task, and limited-data research paradigm (AI 1.0) to a multi-modal, multi-task, massive data, and super-large model research paradigm (AI 2.0). However, how to develop LSF-Model in the field of PHM is still inconclusive. In order to promote the research and application of LSF-Model in the PHM field, this section explains and analyzes how to construct LSF-Models for PHM applications from four aspects.

### 4.1 Large-Scale Datasets for PHM

**4.1.1 Research Status**

Different from the NLP and CV fields, the data in the PHM field is generally high-frequency or low-frequency time-series data collected by various sensors, such as vibration signals [195], sound signals [196], current and voltage [197], temperature [198], pressure [199], etc. In addition, some applications try to use video and image data to achieve equipment health monitoring, such as track defect monitoring [200, 201], defective product identification [202, 203], and equipment crack monitoring [204, 205]. Currently, the PHM community has open-sourced dozens of datasets of different sizes and domains, such as bearing fault datasets [206], bearing degradation datasets [207], gearbox fault datasets [208], aircraft engine degradation datasets [209], three-phase motor fault datasets [210], industrial production monitoring datasets [211], wind turbine monitoring datasets [212], and so on. As the most classic fault diagnosis dataset, the CWRU bearing dataset [195] only contains limited working conditions and four different health categories, and each fault category contains only three fault degrees. The XJTU-SY bearing dataset, released by Wang et al. [207], contains only degradation data of 15 bearings from healthy to failure. The dataset published by Lessmeier et al. [213] includes vibration signal data from 32 bearings and current data from the drive motor. The wafer defect identification dataset released by Wang et al. [211] contains 38,000 wafer data samples, including 38 different types of mixed defect categories. Chao et al. [209] have open-sourced the Turbofan Engine Degradation Simulation dataset, which contains eight sets of data with 128 units and has seven failure modes. It can be seen that the scale of these datasets is very small, and it is challenging to meet the needs of LSF-Models training and optimization. The deep models trained based on these datasets are difficult to understand the nature and laws of PHM data, and it is challenging to have multi-task and zero-shot generalization capabilities. Therefore, constructing a large-scale dataset in the PHM domain is the first step to realizing the LSF-Models.

The advent of Industrial Internet and Internet of Things (IoT) technology [214] has resulted in the installation of numerous sensors on contemporary industrial production equipment and various complex mechanical equipment to facilitate real-time monitoring of the system's various physical quantities for timely detection of anomalies. Therefore, most major enterprises have collected a large amount of industrial data and established corresponding data centers. As an example, China's urban rail trains, with decades of operational experience and accumulated actual operating data, generate massive and comprehensive datasets [215]. A single train can now monitor and record hundreds of variables related to its subsystems, components, and external environment in real-time, such as bearing oil temperature, gearbox noise, and the current and voltage of various systems. Similarly, a rail inspection car can obtain intensive parameters like rail gauge, level, and height over an extended distance. These massive data sets provide valuable information on the real-time status, degradation process, and interdependence of train subsystems and components. Collectively, they lay the groundwork for building the LSF-Model.

Despite the remarkable potential of these industrial data in constructing LSF-Models to solve various PHM tasks, there remains a need for further exploration. Industrial data may comprise a variety of sensor data, such as signals, images, and videos, as well as a significant amount of text information, including maintenance work orders and reports. As such, building LSF-Models to utilize this multi-sensor data effectively presents new challenges. In addition, a single data center's data scale may be limited for building LSF-Model. For example, the training data for GPT 3.0 [58] exceeds 410 billion tokens, while the training data for GPT 3.5 [60] may significantly exceed that of GPT 3.0. To address this issue, joint deep model training and optimization across centers or regions is a viable solution. However, these industrial data may usually involve commercial secrets, so the government and enterprises have established strict data protection regulations, which greatly restrict the sharing and use of industrial data.

### 4.1.2 Solution

Although LSF-Models have shown promising results in NLP and CV, the data in the PHM field differs significantly from that of the former two. Therefore, it is necessary to optimize and improve the key components of LSF-Models according to the unique data characteristics of the PHM field to achieve considerable performance in this domain. This requires exploring novel feature extraction models, self-supervised representation learning algorithms, and multimodal fusion algorithms tailored to the PHM domain. Additionally, in the industrial field, academia has established cutting-edge algorithm design and data analysis capabilities, while the industry has accumulated large-scale industrial monitoring data. Therefore, establishing the school-enterprise joint research center to fully leverage their respective advantages will effectively address existing challenges and significantly promote LSF-Model research in the PHM field.

In addition, the implementation of LSF-Models requires access to large-scale data. Ensuring data privacy protection while utilizing such data is another crucial issue that warrants attention. Unlike NLP and CV fields, where relevant data can be obtained on a large scale from the Internet, PHM datasets are in the hands of equipment operators, and they are often precious and may contain commercial secrets. Moreover, with the increasing focus on data privacy and security, regulatory bodies have introduced new laws to regulate the management and use of data [216]. Consequently, developing solutions that comply with strict privacy protection regulations and solve the challenges of data fragmentation and isolation is necessary. Federated learning [217, 218], a distributed machine learning framework with privacy protection and secure encryption, is a possible solution. It allows decentralized participants to collaborate on machine learning model training without disclosing private data to other participants. Currently, solutions for data privacy protection based on federated learning have been proposed in the PHM field [219-221]. **Fig. 6** shows the multi-vehicle multi-center data privacy security protection architecture based on federated learning. However, most of the existing federated learning algorithms are yet to be deployed in real industrial large-scale data analysis. Therefore, promoting large-scale industrial data analysis and the establishment of LSF-Models requires collaborative efforts from both academia and industry.

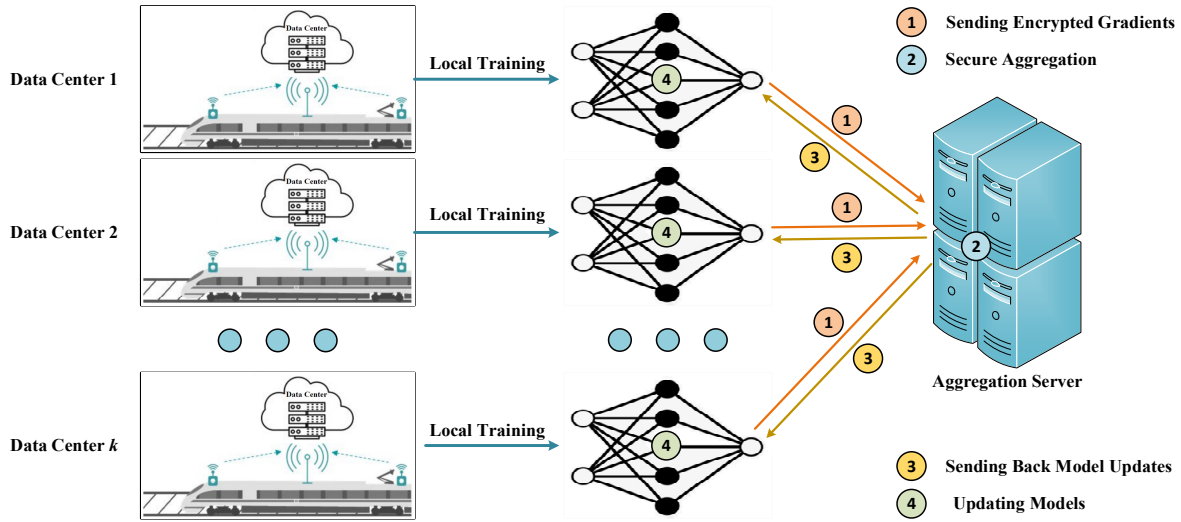

**Fig. 6.** Flowchart of the federated learning framework for multi-vehicle and multi-center data privacy security protection.

## 4.2 Transformer for PHM

### 4.2.1 Research Status

34 The Transformer [75] is a deep learning model that is specialized in modeling long-distance feature correlations, allowing for the modeling of the correlation between any two positions in the signal, irrespective of the actual physical distance between the features. Its efficient long-term dependency modeling capability makes it well-suited for analyzing and processing various sensor data in the PHM field. Therefore, the Transformer has found widespread application in various PHM applications [42, 222-224], achieving impressive performance results. Jin et al. [224] proposed a time series-based Transformer model with superior fault recognition capability compared to traditional CNN and RNN models. Wang et al. [42] proposed a high-speed train wheel wear prediction model based on the Transformer, which leverages the strengths of both the Transformer and CNN to encode global and local information effectively. Fang et al. [222] optimized the Transformer model and proposed a lightweight fault diagnosis framework based on the Transformer, which achieves efficient and accurate fault diagnosis with reduced computational complexity. Despite achieving significant success in the field of PHM, the Transformer model still has some limitations that need to be addressed. Firstly, the Transformer model is primarily designed for processing static input data, such as text, and although it can incorporate timing information through position coding, it does not directly consider timing information [225]. Therefore, it is difficult for Transformer to fully learn the continuous time relationship of data when processing industrial time series data. Secondly, the Transformer model may not perform well for noisy industrial data. In real-world industrial production processes, industrial data containing noise is commonplace, and the Transformer model may not be robust enough to handle such noisy data. Thirdly, as mentioned previously, industrial data often comprise multiple types of sensor data and a significant amount of textual information, posing a new challenge to the design of Transformer architecture. The Transformer model is typically not capable of processing numerous sensor data simultaneously.

### 4.2.2 Solution

Due to its implementation based on the self-attention mechanism, the Transformer model faces challenges in considering the temporal relationship between time series data. The self-attention mechanism is restricted to consider only the relationship between positions in the sequence, lacking the ability to capture the relative time relationship on the time axis. To address this issue, several approaches can be adopted. Firstly, a temporal encoding mechanism can be integrated into the Transformer model to facilitate the direct learning of temporal relationships. Secondly, an effective temporal attention mechanism can be explored to capture temporal dependencies between data within the Transformer architecture. Finally, Transformer-based time series models, such as time series Transformer [226], can be explored considering the characteristics of sensor data. These models need to be specifically designed to process time series data, integrate global time series information, and improve the modeling of time correlation.

Data collected in the industrial field often contains complex noise, unlike NLP and CV fields, where data is generally clean. As a result, most existing deep learning models are ill-equipped to handle the interference of noise or irrelevant information on time series signals. In traditional signal analysis, various methods have been proposed to remove signal noise and extract valuable information. These methods include time and frequency domain filtering [227, 228], Fast Fourier Transform (FFT) [229, 230], wavelet transforms [231, 232], etc. Therefore, the deep integration of Transformer and signal analysis techniques can be a promising solution to this issue. For instance, Rao et al. [233] proposed an FFT-based global filtering layer to replace the Transformer's self-attention mechanism, which resulted in better performance. Similarly, Wang et al. [41] achieved the best performance in a noisy environment by fusing wavelet transform with deep learning models s, and the proposed multi-layer wavelet model is shown in **Fig. 7**. However, how to endow the Transformer with more advanced and effective signal analysis capabilities remains to be further studied.

Finally, to simultaneously process multi-sensor data and industrial text data, it is essential to redesign the Transformer architecture and create the multi-modal Transformers [234-236]. The main idea of the multi-modal Transformer is to embed different types of data (such as text, image, and audio) into the Transformer model, thereby enabling end-to-end modeling when dealing with multi-modal data. Several solutions are feasible, including transformer-based multi-modal embedding, transformer-based cross-modal pre-training, and Transformer-based multi-modal attention. Multi-modal embedding encodes and embeds various sensor data into the Transformer architecture to represent multi-sensor data simultaneously. Cross-modal pre-training utilizes various sensor data to pre-train Transformer to learn valuable information from different data types, and then applies it to downstream multi-sensor data processing. Multi-modal attention is a common solution that achieves the adaptive fusion of multi-sensor information by utilizing attention weights.

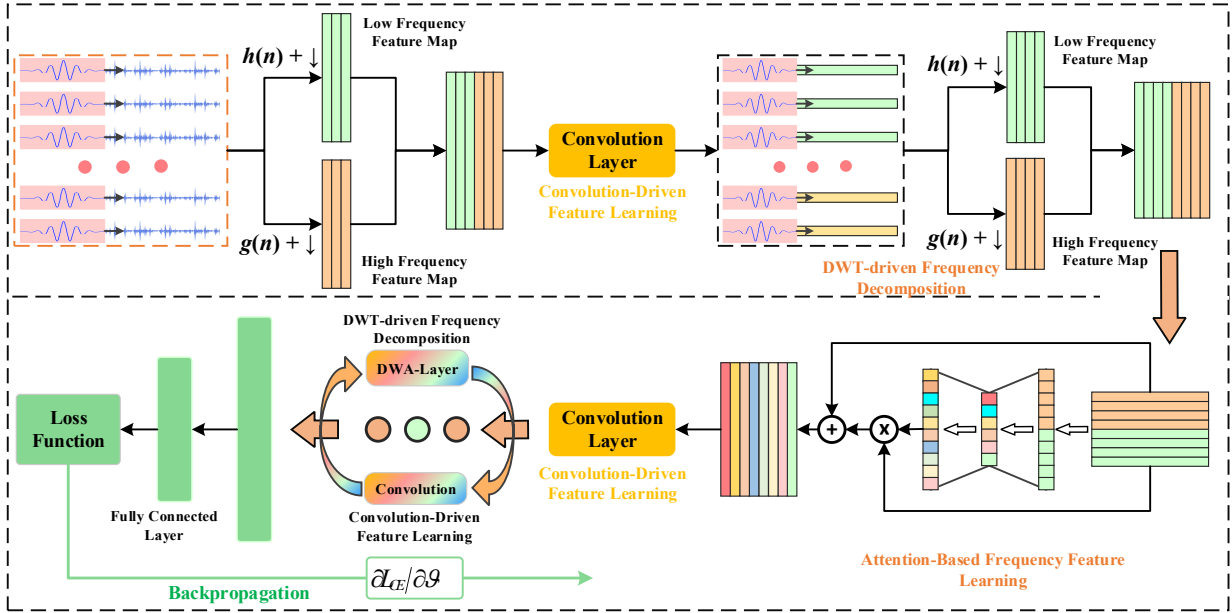

**Fig. 7.** The basic architecture of the multi-layer wavelet model, which deeply integrates the advantages of wavelet transform and CNN, makes it a solid anti-noise ability (image from [41]).

## 4.3 Self-Supervised Learning for PHM

### 4.3.1 Research Status

As previously mentioned, a significant amount of sensor monitoring data is gathered during industrial production and equipment operation. However, academia and industry struggle to fully utilize this vast amount of operational data to build PHM models. The primary challenges are the unavailability of labeled data, the presence of noise, and the large data volume. As a result, it is difficult for existing deep models to effectively extract useful information from such data. Self-supervised learning [70, 124], which can automatically learn valuable feature representations from a large amount of unlabeled data, has emerged as a core algorithm for building PHM foundation models. Currently, self-supervised learning has been researched and applied in the field of PHM [237-241]. For example, Zhang et al. [239] developed a self-supervised algorithm that incorporates prior knowledge and has demonstrated good fault identification capabilities on small labeled datasets. Wang et al. [49] proposed a novel pretext task for self-supervised representation training, which has effectively extracted valuable information from unlabeled signals. Meanwhile, Ding et al. [237] developed a pre-training algorithm based on contrastive learning, which has shown promise in identifying early failures of bearings.

Although self-supervised learning has shown great potential in the field of PHM, it requires the design of effective pretext tasks to learn useful feature representations. Current pretext tasks may not capture fault- and health-related information adequately, particularly in complex industrial systems. Furthermore, self-supervised learning models may be sensitive to noise, which could impact the learned feature representation in the case of poor data quality, thereby reducing the performance of fault diagnosis and health management. Additionally, existing self-supervised learning algorithms have difficulty handling real-world industrial data containing multiple sensor modalities. Therefore, developing effective self-supervised representation learning algorithms that can handle such industrial data is a critical area of focus.

### 4.3.2 Solution

The fundamental aim of researching LSF-Models is to attain few-shot or zero-shot generalization. Therefore, the primary question that needs to be solved when designing models and algorithms is: What kind of models and algorithms can achieve zero-shot generalization [64]? Autoencoder-like architectures [53], specifically those pertaining to signal reconstruction, signal completion, signal denoising, etc., constitute an effective solution. However, their feature representation process lacks a direct correlation with health information. Hence, while developing pretext tasks, it is essential to ensure maximum relevance with equipment health.

In the domain of PHM, the frequency domain is a more effective method for reflecting the fault and health status information of equipment than the time domain. Moreover, analysis from the frequency domain level better tackles noise interference and enhances model noise robustness. Thus, it is imperative to comprehensively consider signal frequency domain information while constructing self-supervised learning pretext tasks. There exist several potential solutions for this. 1) Utilize both time-domain and frequency-domain features concurrently in self-supervised learning algorithms, which can be fused into a unified feature representation. This involves using the original time-domain data and its corresponding frequency-domain data as input to train a multimodal self-supervised learning model to learn time-domain and frequency-domain features. 2) Develop a deep frequency space learning model, such as wavelet-based CNN [41, 46, 47], which can directly learn frequency features and capture frequency information in the data. 3) Construct pretext tasks that rely on the frequency domain data of the signal, such as reconstructing frequency domain information. 4) Create contrastive learning algorithms based on time-frequency consistency [242] of signals for effective self-supervised training on time-series data. 5) The correlation of multi-sensor data can be used to construct pretext tasks. For instance, a part of sensor data can be used to predict the value of other sensor data, to induce the model to learn the direct interdependence of multi-sensors and facilitate the feature representation learning of industrial multi-sensor data. Finally, different self-supervised learning algorithms can represent different feature information of the data. Combining multiple self-supervised learning algorithms can help the model learn a more diverse set of feature representations, improving performance in downstream tasks. Therefore, using various algorithms to comprehensively pre-train the model in practice is an excellent solution.

### 4.4 Multi-Modal Fusion for PHM

### 4.4.1 Research Status

In the PHM field, the primary data source is discrete time-series data obtained from various sensors, including vibration signals, current and voltage, temperature, and pressure. Furthermore, certain applications attempt to utilize video and image data for the purpose of equipment health monitoring, such as track defect monitoring, defective product identification, and equipment crack monitoring. Data within the industrial domain, may encompass numerous sensor data types (e.g., signals, images, videos, etc.) and a substantial amount of textual information (e.g., maintenance work orders, maintenance reports, etc.). Therefore, compared to the fields of NLP and CV, the PHM field places greater emphasis on information fusion of multi-sensor data to achieve a more comprehensive understanding of equipment health status. Currently, several works have explored the fusion of multi-sensor information [243-248]. For example, Yang et al. [243] presented a multi-sensor and multi-scale fusion model based on the kurtosis weighting algorithm and pyramid principle, which demonstrated strong performance in bearing health identification. Long et al. [245] employed the Hilbert transform and FFT to extract valuable frequency information from multi-sensor signals and then developed an attention-based model for fault identification. Kumar et al. [248] simultaneously considered vibration and sound signals, utilizing wavelet transform as a feature extractor, and subsequently applied machine learning methods for fault diagnosis.

However, most existing datasets only contain data from one to three sensors, which is far from meeting the needs of practical industrial applications. Moreover, as the number of sensors increases, the data generated by different sensors may have different distributions, scales, and signal-to-noise ratios. Appropriate preprocessing methods and fusion strategies are crucial to the performance of PHM models. In addition, in the industrial field, multimodal fusion algorithms must consider the fusion of various industrial sensor information (such as signals, images, text, etc.). Nevertheless, existing research lacks sufficient solutions and optimization strategies for this challenge.

### 4.4.2 Solution

Effective multi-sensor data fusion in PHM requires comprehensive efforts in data fusion, feature fusion, and model fusion. The following solutions can enhance the multi-sensor fusion ability of models:

1) Explore new methods to integrate data from multiple sensors or sources into a unified data representation that considers the unique characteristics of each sensor. Such a data fusion approach can improve model performance and stability when dealing with data of different modalities.

2) Construct appropriate attention mechanisms for specific tasks and data characteristics to achieve efficient multi-sensor correlation modeling and fuse valuable features from sensors. This mechanism enables models to capture the correlation among sensors and highlight relevant features.

3) Optimize the structure and parameters of models to better adapt to the characteristics of different sensor data. Moreover, for different sensor data, building distinct models and then employ model fusion algorithms to establish connections among these models and achieve multi-sensor information fusion.

By tailoring models to the specific characteristics of different sensors, performance can be improved, and more accurate predictions can be made.

4) The graph neural network (GNN)-based multimodal fusion algorithm [249-251] can be studied to model complex relationships among multimodal data. This method uses GNN to construct the graph structure of multimodal data and takes the information of different sensors or data types as node input to realize the fusion of multimodal data.

Finally, the solutions given above are all general solutions, which need to fully combine feature extraction models and self-supervised algorithms for specific data and applications to achieve the optimal performance of multimodal data fusion.

# 5. Challenges and Future Roadmaps

In the previous section, we discussed the technical details and feasible solutions for building LSF-Models in the PHM domain. This section attempts to discuss the challenges, roadmaps, and prospects of these models from a broader, more global perspective. By doing so, we can better understand the bigger picture and identify areas for improvement and future research in the field of PHM.

## 5.1 Challenges of Large-Scale Foundation Models in PHM

**Fig. 8** shows the challenges faced by LSF-Models in the field of PHM, which mainly include seven aspects of datasets, security, credibility, transferability, advanced cognition, interpretability, and edge device deployment. They are described in detail below.

### 5.1.1 Large-Scale Dataset

Enterprises are reluctant to reveal their commercially sensitive information, including performance parameters of vital equipment, maintenance records, and production process failures, to their competitors. This information can be exploited by competitors to enhance competitiveness or damage the corporate reputation. Industrial production data and mechanical equipment operation monitoring data used for PHM applications will inevitably involve this commercially sensitive information. Consequently, enterprises have established rigorous data protection measures, resulting in a severe scarcity of large-scale publicly accessible industrial data in the PHM domain. Moreover, modern industrial equipment often functions in intricate and unbounded settings, leading to the collection of data that is frequently fraught with noise, missing data, distortion, abnormal disturbance, and other issues. Additionally, the diverse array of industrial sensors deployed results in the acquisition of varied data types, necessitating the processing of dissimilar modal data. As a result, the construction of effective large-scale PHM models using such data is challenging.

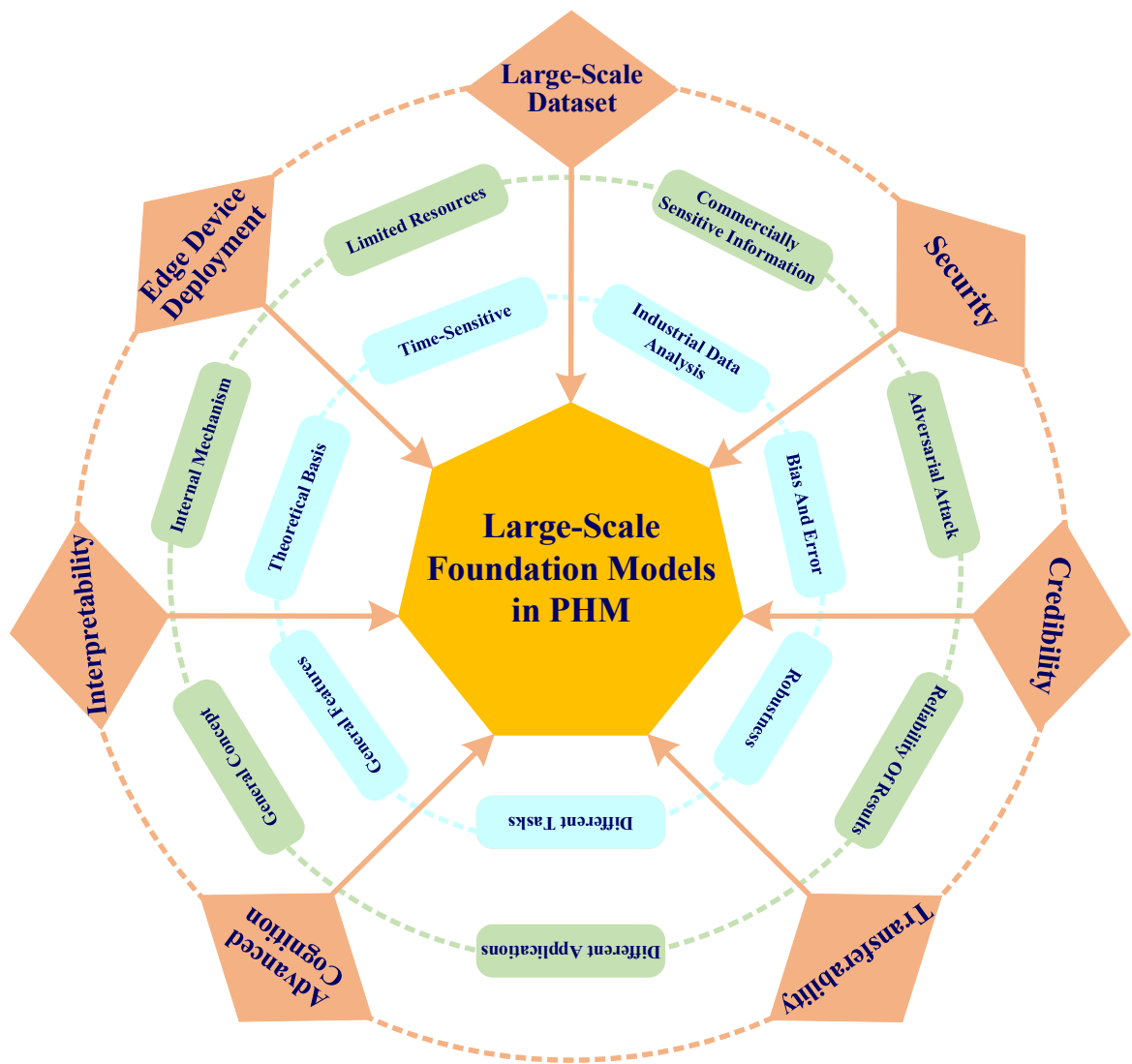

**Fig. 8.** Challenges faced by LSF-Models in the field of PHM.

### 5.1.2 Security

Even though the LSF-Models have been trained on massive data sets, they remain vulnerable to security threats, such as adversarial attacks [252]. Adversarial attacks can interfere with the output of PHM models through well-designed noises, causing the models to produce wrong results. This is unacceptable in critical PHM applications, such as the safety monitoring of nuclear power plants. In addition to direct attack, unintentional bias or errors may be introduced during the training process on large-scale data, which may cause the model to output biased and even wrong results. In this case, the PHM model may have security risks and be challenging to troubleshoot, because specific biases and erroneous results only occasionally occur in specific scenarios. In addition, since LSF-Models require massive computing resources for training and inference, they are also vulnerable to attacks targeting computing resources, such as DDoS attacks [253, 254] and malware attacks [255, 256].

### 5.1.3 Credibility

The credibility [257] of the LSF-Model represents the user's confidence in the accuracy and consistency of its output. Additionally, the model must withstand disturbances such as bias, noise, and input data perturbations while maintaining prediction accuracy and stability. Currently, the credibility of LSF-Model has significantly improved compared to traditional deep learning models. However, in the PHM domain, complex working conditions, diverse background noises, and varied old equipment lead to considerable data distribution disparities. Consequently, enhancing LSF-Model credibility remains a significant challenge.

### 5.1.4 Transferability

The transferability of the LSF-Model refers to whether the model can be well-migrated and adapted to different tasks and datasets. In the PHM domain, different tasks (e.g., anomaly detection, fault prediction, RUL estimation) are required for different scenarios. However, the nature and implementation of these PHM tasks are quite different, and how to achieve considerable transferability and adaptability across various tasks is a challenge. In addition, different mechanical equipment has different properties; for example, bearings, gearboxes, and crankshafts are all core rotating equipment, but their failure characteristics and properties vary greatly. Another challenge is how to deal with these differences to make LSF-Models have strong transferability and adaptability under various applications.

### 5.1.5 Advanced Cognition

Current research on large-scale language and visual models has demonstrated that LSF-Models possess advanced cognitive abilities and can learn general concepts to some extent [62, 64]. However, it remains unclear how these models acquire these abilities and how they can be explicitly enhanced. Moreover, PHM data exhibit significant randomness, posing a challenge for models to generalize high-level cognition from this data. In contrast to text and visual data, which possess regularity, time series data's diversity makes it harder for the model to extract generalized feature representations.

### 5.1.6 Interpretability

Interpretability [258-261] means that the decisions and outputs of the model can be interpreted and understood, and transparency means that the internal structure and operation of the model can be understood and reviewed by humans. However, with the development of LSF-Models, the inner workings and operating mechanisms of these models become more challenging to explain and understand, and the transparency becomes less and less. Additionally, LSF-Models may capture unpredictable data relationships and noise during training, which could lead to significant security risks. In the field of PHM [41, 262, 263], interpretability and transparency are of significant importance since models lacking transparency are challenging to trust in domains such as nuclear energy and aerospace. However, the complexity of LSF-Models makes studying their interpretability and transparency extremely difficult.

### 5.1.7 Edge Device Deployment

LSF-Models have massive parameters and complex structures, necessitating significant computing resources for training and inference. However, in the PHM field, it is often necessary to deploy monitoring models to edge computing devices or mobile terminals to meet real-time and security requirements. However, as existing edge computing devices generally lack the capability to execute LSF-Models, it poses a significant challenge to implement the deployment of such models without compromising their performance. Therefore, it is imperative to devise novel techniques to address this issue and facilitate the deployment of LSF-Models on edge computing devices [264-266] while preserving their performance.

## 5.2  Roadmaps of Large-Scale Foundation Models in PHM

**Fig. 9** shows the future roadmaps of LSF-Models in the field of PHM. According to the challenges faced by LSF-Models, this section explores how to solve these challenges and elaborates on the roadmap for the future. They are described in detail below.

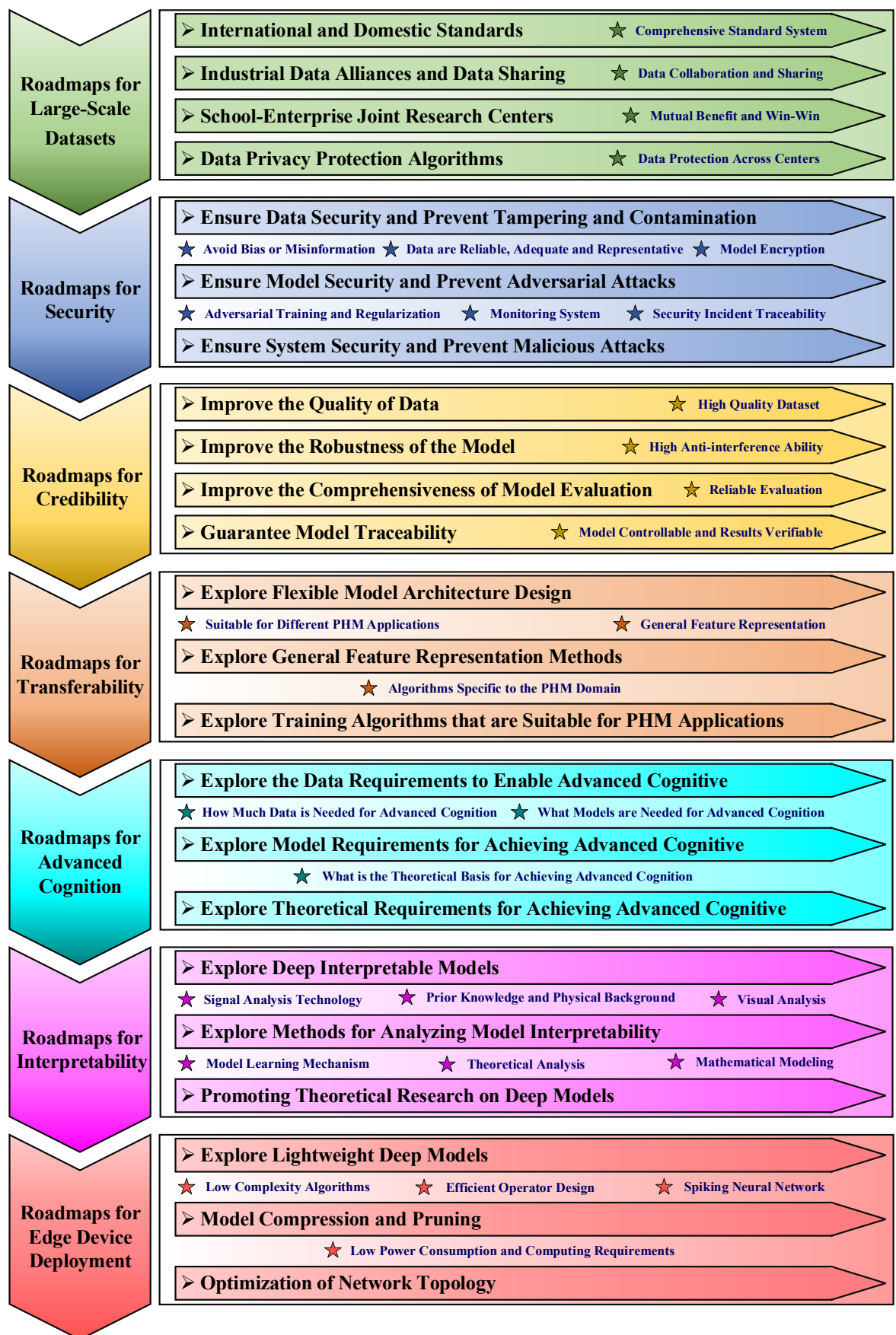

**Fig. 9.** Roadmaps of LSF-Models in the field of PHM.

**5.2.1 Roadmaps for Large-Scale Datasets**

The core solution to constructing large-scale datasets in the PHM field is to eliminate the concerns of enterprises about data privacy and security, especially to enable all enterprises to obtain benefits, so as to achieve mutual benefit and win-win results. To this end, the analysis and utilization of large-scale datasets in the PHM field can be advanced through the following approaches.

1) **Construction of International and Domestic Standards.** First, under the joint promotion of academia and industry, build a complete set of the standard system, including data collection and storage standards, etc. Then, use these standards to standardize the industry's data collection methods, data collection objects, data collection frequency, data collection quality, data storage format, and data privacy protection requirements.

2) **Build Industrial Data Alliances to Realize Data Sharing.** Under the promotion of the government or industry associations, establish an industrial data alliance to realize industry data sharing while protecting the business secrets of each enterprise and achieve mutual benefit and win-win results. Models built using large-scale industry data can feed back participating companies to improve their products' quality, reliability, and safety.

3) **Promote the Construction of School-Enterprise Joint Research Centers.** In the industrial field, the academia has mastered cutting-edge algorithm design and data analysis capabilities, and the industry has mastered large-scale industrial data. Therefore, establishing school-enterprise joint research centers can synergistically leverage their respective strengths, facilitating the advancement of research on LSF-Models in the PHM field.

4) **Explore Algorithms with Data Privacy Protection.** Efficient cross-data center large-scale deep learning model training and optimization algorithms should be explored while protecting data privacy. To address concerns over data privacy and security among enterprises, techniques such as data desensitization, encryption, and other means can be employed to facilitate data sharing.

**5.2.2 Roadmaps for Security**

The security of LSF-Models can be mainly divided into data security, model security, and system security. To this end, the security of LSF-Model can be improved from the following aspects.

1) **Ensure Data Security.** Ensuring data security requires avoiding malicious tampering or contamination of the data and avoiding introducing bias or wrong information to the model. Therefore, ensuring that the model's training data source is reliable, high-quality, sufficient, and representative is necessary. In addition, a data security protection mechanism needs to be established to avoid malicious data leakage and tampering.

2) **Ensure Model Security.** Explore practical adversarial training and regularization methods [267-269] for LSF-Models to improve the model's ability to resist malicious disturbances and noise. In addition, enhance the security and privacy of the model, such as model encryption [270-272], to prevent the model from being stolen or reversed.

3) **Ensure System Security.** For the LSF-Model-based PHM system, establish a monitoring system to detect abnormal operations and attack behaviors in time and record operation and audit logs simultaneously to investigate and trace security incidents.

### 5.2.3 Roadmaps for Credibility

Improving the credibility of LSF-Models is a comprehensive and complex issue, which can be optimized and improved from the following aspects.

1) **Improve the Quality of Data.** Data quality forms the basis for model training, and ensuring data quality is essential for enhancing the model's credibility. Therefore, a series of measures need to be taken to ensure data quality, including data cleaning, removing outliers, data enhancement, etc.

2) **Improve the Robustness of the Model.** Explore new solutions to improve the robustness and anti-interference ability of the model, such as improving the performance of the model in noisy [9, 10, 273-276], distribution shift [277-281] and out-of-distribution scenarios [282-284].

3) **Improve the Comprehensiveness of Model Evaluation.** The comprehensiveness and fairness of evaluation are crucial in improving the model's credibility. To achieve this, it is necessary to comprehensively evaluate models using multiple evaluation indicators and multiple datasets while avoiding problems such as model bias for specific scenarios.

4) **Guarantee Model Traceability.** In order to improve the credibility of the model, it is necessary to fully record and trace the development, training, testing, deployment, and other processes of the model. This can ensure that the process of the model is controllable and the results can be verified, thereby improving the credibility of the model.

### 5.2.4 Roadmaps for Transferability

The objective of conducting LSF-Models is to advance the concept of a universal model capable of performing all tasks. With this aim in mind, we are dedicated to creating a comprehensive multi-task PHM system. To achieve this goal, it is the top priority to improve the portability and adaptability of foundation models as much as possible. The PHM field encompasses a diverse range of downstream tasks, and its application objects are complex, thus necessitating the exploration of new architecture design approaches, feature learning methods, and training algorithms for real-world industrial scenarios to enhance transferability.

1) **Explore Flexible Model Architecture Design.** To construct a general-purpose foundation model, it is necessary to investigate flexible model architecture designs that can be customized for different types of PHM applications, such as monitoring, diagnosis, prediction, and others.

2) **Explore General Feature Representation Methods.** Extracting valuable features from different data types is the key to realizing general models. These feature representation algorithms need to be able to capture various types of failure modes and also need to be robust to different types of failure data.

3) **Explore Training Algorithms that are Suitable for PHM Applications.** Although a large number of self-supervised training algorithms [49, 69-71, 124, 125] have been proposed in the fields of NLP and CV, only the training algorithms designed based on the characteristics of the PHM field can improve the transferability and adaptability of the foundation model to the greatest extent.

**5.2.5 Roadmaps for Advanced Cognition**

LSF-Models have the potential to acquire advanced cognition and learn general concepts from data using vast amounts of training data and well-designed algorithms. In the field of PHM, LSF-Models with advanced cognitive capabilities can well understand the nature and general concepts of industrial data. As a result, LSF-Models can promptly adapt to varying industrial scenarios and tasks, effectively avoiding the generation of unintelligible outputs. However, the advanced cognitive capabilities displayed by current LSF-Models are inadequate. In addition to scaling up the amount of training data, the following aspects require exploration:

1) **Explore the Data Requirements to Enable Advanced Cognitive Capabilities.** Although some existing LSF-Models have demonstrated limited advanced cognitive abilities, what kind of data and how much data is needed to achieve such abilities is unknown. In addition, how to make full use of data information to obtain considerable cognitive ability on relatively small data sets is also a topic worthy of research.

2) **Explore Model Requirements for Achieving Advanced Cognitive Capabilities.** Most existing LSF-Models adopt Transformer architecture, but it has limitations. Hence, exploring new Transformer architectures or designing novel model architectures to enable LSF-Models to exhibit better cognitive capabilities are areas that warrant exploration.

3) **Explore the Theoretical Requirements for Achieving Advanced Cognitive Capabilities.** The realization of advanced cognitive abilities cannot rely solely on a considerable amount of experimental analysis; it also requires appropriate theories to provide guiding solutions. Therefore, exploring the reasons and mechanisms behind the acquisition of advanced cognitive abilities by LSF-Models will significantly advance the field.

**5.2.6 Roadmaps for Interpretability**

Deep learning has been criticized for its lack of interpretability and transparency, and this deficiency is more evident in LSF-Models. Improving the interpretability and transparency of these models would enhance their credibility, controllability, and reliability, making them more suitable for application in critical industrial fields such as aerospace and nuclear energy. Future research can be explored from the following three aspects to address this issue.

1) **Explore Deep Interpretable Models.** In the field of PHM, a feasible solution is to realize a deep model with high interpretability by means of the theoretical basis of signal analysis methods [41, 46, 47, 263]. Moreover, incorporating prior knowledge or physical background information into the model can also enhance its interpretability significantly.

2) **Explore Methods for Analyzing Model Interpretability.** According to the characteristics of deep models, practical methods for analyzing model interpretability need to be explored. For example, by using the visualization method [41, 48, 285, 286] to analyze the model's internal structural parameters and the signal analysis method to analyze the model's filter [287] to understand which features the model has learned.

3) **Promoting Theoretical Research on Deep Models.** This requires an in-depth analysis of the principles of the foundation model, exploring its essential attributes, characteristics, advantages, and disadvantages and revealing its internal mechanism and laws through theoretical analysis [288-290]. By establishing a mathematical model of the foundation model and deriving its convergence, generalization, and robustness properties, for example, the internal mechanism of deep models can be revealed.

### 5.2.7 Roadmaps for Edge Device Deployment

Despite the impressive performance of LSF-Models, their computational demands significantly limit their applicability. In order to promote the application of LSF-Models in the field of PHM, the following aspects can be explored to make them lightweight.

1) **Explore Lightweight Deep Models.** By exploring the design method of the lightweight deep model [291, 292] and developing new low-complexity algorithms and operator design paradigms, the computational load and resource usage of the model can be reduced.

2) **Model Compression and Pruning.** Model compression [293, 294] and pruning [295, 296] techniques can be explored to reduce the model's size. Efficient compression methods such as weight sharing, low-rank decomposition, and knowledge distillation can be employed. Additionally, redundant neurons or connections can be eliminated using efficient pruning methods.

3) **Optimization of Network Topology.** Explore new network topology design schemes, such as brain-inspired spiking neural networks [297, 298], to achieve deep network models with low power consumption and computational requirements.

## 5.3 Prospect: Universal PHM Platform

Traditional AI-based models for PHM typically require the development of unique data processing and model optimization algorithms for each subtask to achieve good performance in specific scenarios. For example, it is necessary to build independent fault diagnosis and RUL estimation models for bearings. However, complex industrial equipment may contain hundreds of core components, and it is a vast project to build corresponding PHM models for all components. In addition, this application paradigm cannot fully utilize the multimodal interrelated data of industrial equipment and cannot comprehensively characterize the health status of industrial equipment. Moreover, the various subtask models constructed are time-consuming and laborious and cannot guarantee that the model has good generalization. As mentioned earlier, the fields of NLP and CV, such as ChatGPT [62] and SAM [64], have demonstrated the fantastic performance of LSF-Models in cross-modal multi-task scenarios, and they have realized the idea of a model for all tasks to a certain extent. Therefore, inspired by this, it becomes possible to implement universal PHM platforms to realize one model for all PHM tasks. This research concept will completely revolutionize the research paradigm in the PHM field, and the realization of universal PHM platforms will significantly promote the comprehensive intelligence of the PHM field.

**Fig. 10** presents a schematic diagram of the universal PHM platform for high-speed trains. The core of the universal PHM platform is the large-scale cross-modal foundation model. This model receives various data from high-speed trains, such as sensor data (including signals, images, and videos), maintenance work orders and records, as well as expert experience and knowledge, as input. Subsequently, it conducts comprehensive assessment and monitoring of the health status of various subsystems and core components of high-speed trains. Specifically, the cross-modal foundation model is constructed by employing the multi-modal fusion algorithm and the self-supervised feature representation algorithm, which enables it to understand and learn the universal and high-quality data features of multi-modal train data. The zero/few-shot learning paradigm is subsequently utilized to extend the cross-modal foundation model to various train PHM sub-tasks, thereby achieving platform effects. The ultimate objective of the cross-modal foundation model is to offer potent and generalizable data mining and understanding tools for PHM tasks, such as health monitoring, fault prediction, anomaly detection, RUL estimation, maintenance planning, and health management, by learning a vast amount of train operation data. Even for novel objects and data that were not present during training, the cross-modal foundation model can still provide a considerable feature mining capacity. Therefore, the proposed universal PHM platform is expected to become a core tool in the PHM field, playing a critical role in improving the efficiency of industrial equipment health management and reducing maintenance costs.

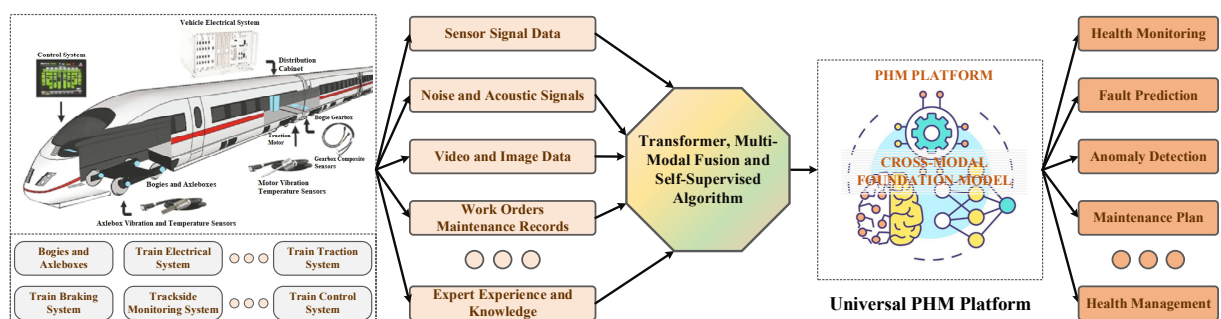

**Fig. 10.** The schematic diagram of the universal PHM platform, taking high-speed train PHM application as an example.

## 6. Conclusions

At present, the research of deep learning is undergoing a new revolution, that is, the research paradigm (AI 1.0) of single mode, single task, and limited data is developing rapidly to the research paradigm (AI 2.0) of multimodal, multi-task, massive data, and super-large model. AI 2.0 focuses on developing LSF-Models with cross-domain knowledge, which can show strong generalization and multi-task ability after training on massive datasets. To this end, this paper provides a comprehensive overview of the three major technical points of LSF-Models and analyzes the research status of LSF-Models in NLP and CV. The literature review shows a severe lack of research on LSF-Models in the PHM field, and there is no feasible solution for how to build LSF-Models for PHM applications. Therefore, this paper comprehensively answers how to build the LSF-Models suitable for the PHM field from four aspects: dataset, deep model, learning algorithm, and data fusion. Finally, this paper attempts to discuss the challenges and roadmaps of LSF-Models from a broader, more global perspective. Overall, this survey systematically introduces LSF-Models and their research status, challenges, solutions, roadmaps, and prospects in the PHM field, which is expected to provide valuable guidance for future research in this field.